\documentclass[10pt,twocolumn,letterpaper]{article}

\usepackage{iccv}
\usepackage{times}
\usepackage{epsfig}
\usepackage{booktabs}
\usepackage{caption}
\usepackage{graphicx, amsmath, amssymb, subcaption, multirow, overpic, textpos}
\usepackage[table]{xcolor}
\usepackage[english]{babel}
\usepackage{xcolor}
\usepackage{soul}
\usepackage{bbding}

\usepackage[pagebackref=false, breaklinks=true, colorlinks,
            citecolor=citecolor, linkcolor=linkcolor, bookmarks=false]{hyperref}
\definecolor{citecolor}{HTML}{0071BC}
\definecolor{linkcolor}{HTML}{ED1C24}
\definecolor{Ins_enc}{HTML}{4672C4}
\usepackage[capitalize]{cleveref}
\crefname{section}{Sec.}{Secs.}
\Crefname{section}{Section}{Sections}
\Crefname{table}{Table}{Tables}
\crefname{table}{Tab.}{Tabs.}

\newlength\savewidth\newcommand\shline{\noalign{\global\savewidth\arrayrulewidth
  \global\arrayrulewidth 1pt}\hline\noalign{\global\arrayrulewidth\savewidth}}
\newcommand{\tablestyle}[2]{\setlength{\tabcolsep}{#1}\renewcommand{\arraystretch}{#2}\centering\footnotesize}
\renewcommand{\paragraph}[1]{\vspace{1.25mm}\noindent\textbf{#1}}

\newcolumntype{x}[1]{>{\centering\arraybackslash}p{#1pt}}
\newcolumntype{y}[1]{>{\raggedright\arraybackslash}p{#1pt}}
\newcolumntype{z}[1]{>{\raggedleft\arraybackslash}p{#1pt}}

\newcommand{\app}{\raise.17ex\hbox{$\scriptstyle\sim$}}

\definecolor{deemph}{gray}{0.6}

\definecolor{baselinecolor}{gray}{.9}
\definecolor{deltacolor}{gray}{.45}
\newcommand{\baseline}[1]{\cellcolor{baselinecolor}{#1}}

\definecolor{airforceblue}{rgb}{1.00, 0.501, 0.01}
\definecolor{egreen}{rgb}{0, 0.69, 0.314}

 \iccvfinalcopy 

\ificcvfinal\fi

\begin{document}

\title{Agglomerative Transformer for Human-Object Interaction Detection}

\author{%
 Danyang Tu$^1$,
 Wei Sun$^1$,
 Guangtao Zhai$^1$,
 Wei Shen$^2$\\
 $^1$Institute of Image Communication and Network Engineering, Shanghai Jiao Tong University\\
 $^2$MoE Key Lab of Artificial Intelligence, AI Institute, Shanghai Jiao Tong University\\
 {\tt\small \{danyangtu, sunguwei, zhaiguangtao, wei.shen\}@sjtu.edu.cn} \\
}
\maketitle
\ificcvfinal\thispagestyle{empty}\fi

\begin{abstract}
   We propose an \textbf{ag}glomerative Transform\textbf{er} (AGER) that enables Transformer-based human-object interaction (HOI) detectors to flexibly exploit extra instance-level cues in a single-stage and end-to-end manner for the first time. AGER acquires instance tokens by dynamically clustering patch tokens and aligning cluster centers to instances with textual guidance, thus enjoying two benefits: 1) Integrality: each instance token is encouraged to contain all discriminative feature regions of an instance, which demonstrates a significant improvement in the extraction of different instance-level cues and subsequently leads to a new state-of-the-art performance of HOI detection with 36.75 mAP on HICO-Det. 2) Efficiency: the dynamical clustering mechanism allows AGER to generate instance tokens jointly with the feature learning of the Transformer encoder, eliminating the need of an additional object detector or instance decoder in prior methods, thus allowing the extraction of desirable extra cues for HOI detection in a single-stage and end-to-end pipeline. Concretely, AGER reduces GFLOPs by $8.5\%$ and improves FPS by $36\%$, even compared to a vanilla DETR-like pipeline without extra cue extraction. 
   The code will be available at~\href{https://github.com/six6607/AGER.git}{https://github.com/six6607/AGER.git}.
\end{abstract}

\section{Introduction}

Human-object interaction (HOI) detection aims at understanding human activities at a fine-grained level. It involves both the localization of interacted human-object pairs and the recognition of their interactions, where the latter poses the major challenges as a higher-level vision task~\cite{interactions}. 

Since interactions describe the relations between different instances (\emph{i.e.}, humans and objects), instance-level cues (\emph{e.g.}, human pose and gaze) are unanimously recognized as pivotal to discriminating subtle visual differences between similar relation patterns in interaction recognition. However, extracting these instance-level cues intuitively indicates a multi-stage pipeline, where an off-the-shelf object detector is essential to generate instance proposals firstly~\cite{no-frills, wan2019pose, li2020detailed, drg, xu2019interact,zhang2022exploring}. Such a paradigm struggles in proposal generation, yielding less competitive performance in model efficiency. In this work, we seek to explore a \emph{single-stage} Transformer-based HOI detector to flexibly and efficiently exploit extra instance-level cues, thus continuing their success in HOI detection.

\begin{figure}[t]
\centering
\includegraphics[width=0.97\linewidth ]{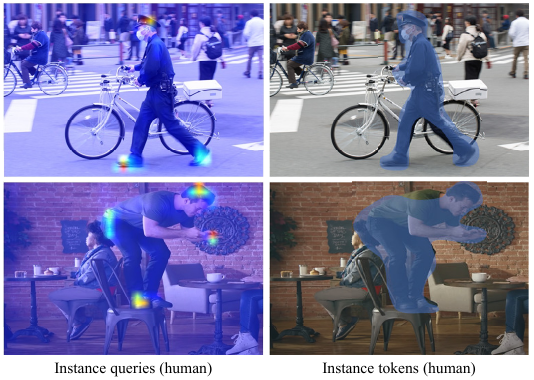}
\captionsetup{size=small}
\caption{\textbf{Instance queries vs. instance tokens}. Instance queries typically attend to instance parts, while our instance tokens are encouraged to contain integral discriminative regions of instances. More examples are presented in supplementary materials.} 
\label{img:example}
\end{figure}
The challenge stems from task-bias, \emph{i.e.}, different tasks have different preferences of discriminative feature regions~\cite{zhou2016learning}. Gaze tracking, for example, prefers the discriminative regions of human heads~\cite{tu2022end}, whereas pose estimation favours holistic human body contexts~\cite{li2021pose}. Therefore, the crux of building a single-stage pipeline lies in a proper design of information carrier, which need to ensure the integrality of instance-level representations (IRs), \emph{i.e.}, containing all discriminative regions of an instance to satisfy the diverse region preferences of different tasks. However, most popular Transformer-based detectors deal with local patches, neglecting the integrality of different instances.

Some previous methods partially tackled the above challenge.  STIP~\cite{zhang2022exploring} employs an additional DETR detector to generate instance proposals, which yet suffers from the low efficiency of the multistage pipeline. Several works~\cite{liao2022gen, chen2021reformulating, hotr} propose to use an additional query-based instance decoder to extract instance queries individually.  Despite being ingenious, these queries are task-driven and learned to highlight only the most distinguishable feature regions preferred by a given task, as verified by the sparsity of learned attention map~\cite{deformableDETR}. As shown in Fig.~\ref{img:example}, the object detection driven human queries in existing methods typically contain only instance extremities, which likewise fails to guarantee the integrality of IRs, limiting its adaptability to other tasks (\emph{e.g.}, pose estimation) due to task bias (Sec.~\ref{sec:clustering}). Although joint multitask learning can partially alleviate the sparsity of instance queries, it introduces unexpected ambiguities and makes the model fitting harder~\cite{xu2022mtformer}.

In this paper, we present AGER, short for \underline{\textbf{AG}}glomerative  Transforme\underline{\textbf{ER}}, a new framework that improves prior methods by proposing instance tokens, handling the above-mentioned challenges favorably. Specifically, we formulate tokenization as a text-guided dynamic clustering process, which progressively agglomerates semantic-related patch tokens (\emph{i.e.}, belonging to the same instance) to enable the emergence of instance tokens through feature learning. Being decoupled from downstream tasks, the clustering mechanism encourages instance tokens to ensure the integrality of extracted IRs (Fig.~\ref{img:example}) and eliminate task bias, thus allowing a flexible extraction of different instance-level cues for HOI detection. Despite being conceptually simple, instance tokens have some striking impacts. Unlike instance proposals being regular rectangles, the instance tokens are generated over irregularly shaped regions that are aligned to different instances with arbitrary shapes (Fig.~\ref{img:example}), thus being more expressive. With this, AGER already outperforms QPIC~\cite{qpic} by $\mathbf{10.6}\%$ mAP even without involving any extra cues (Sec.~\ref{sec:eff}). Additionally, compared to instance queries, instance tokens demonstrate a significant precision improvement in cue extraction (Fig.~\ref{fig:cues}), leading to a new state-of-the-art performance of HOI detection on HICO-Det~\cite{relate1} with $\mathbf{36.75}$ mAP. Of particular interest, the dynamical clustering mechanism can be seamlessly integrated with Transformer encoder, dispensing with additional object detectors or instance decoders and showing an impressive efficiency. Concretely, taking as input an image with size of $640 \times 640$, AGER reduces GFLOPs by $\mathbf{8.5}\%$ and improves FPS by $\mathbf{36.0}\%$ even compared to QPIC that built on an vanilla DETR-like Transformer pipeline (Sec.~\ref{sec:eff}), and the relative efficiency gaps become more evident as the image resolution grows (Fig~\ref{fig:m_size}).

\begin{figure*}[t]
\centering
   \includegraphics[width=\linewidth]{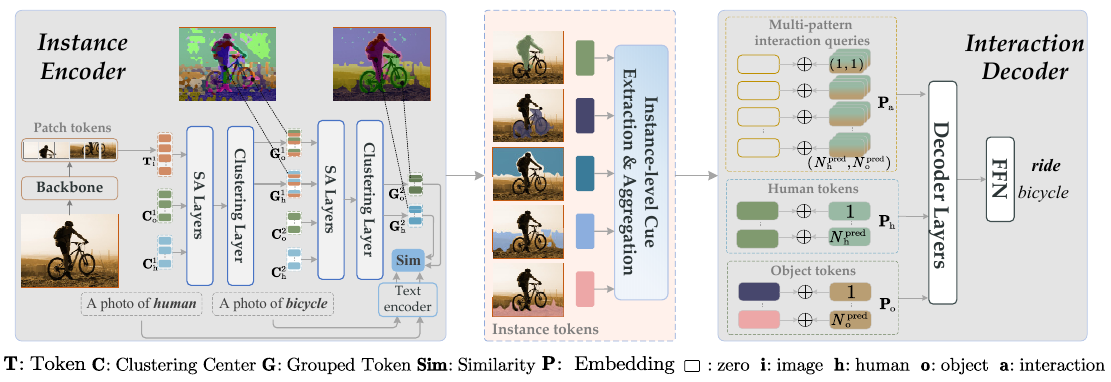}
\captionsetup{size=small}
\caption{\textbf{Architecture of AGER.} AGER performs tokenization as a text-guided dynamic clustering process in the instance encoder, dispensing with any additional object detector or instance decoder, which outputs instance tokens that encourage the integrality of instance-level representations. This property enables the extraction of different instance-level cues in a \emph{single-stage} pipeline. Finally, a new interaction decoder leverages these desirable cues to recognize interactions in a multi-pattern manner.} 
\label{img:whole_model}
\end{figure*}

\section{Related Work}
\label{sec:ret_wok}
Modern HOI detection methods are built on three different information carriers of IRs, \emph{i.e.}, instance proposals, points and instance queries, which show different effects on the utilization of instance-level cues.

\vspace{1mm}
\noindent
\textbf{Instance proposals} dominated CNN-based HOI detection approaches for almost the entire era ~\cite{relate1,drg,ican,interactions,no-frills,zhi_eccv2020,uniondet,kim2020detecting,
li2020detailed,li2019transferable,lin2020action,liu2020amplifying,qi2018learning,ulutan2020vsgnet,wan2019pose,wang2020contextual,wang2019deep,xu2019interact,yang2020graph,zhong2021polysemy,zhou2019relation,zhou2020cascaded}. These methods conventionally shared a two-stage pipeline, employing an object detector~\cite{ren2015faster, he2017mask} to generate instance proposals in the first stage. Then, the human and object proposals are processed separately to extract various instance-level cues, such as human pose~\cite{no-frills,wan2019pose,li2020detailed}, human parsing~\cite{liu2020amplifying}, spatial configurations~\cite{drg}, human gaze~\cite{xu2019interact}, object labels~\cite{zhong2021polysemy}, among others. Along with visual features of appearance, these auxiliary cues were leveraged either individually or conjunctively to further reason about the interactions. Although a fine-selected proposal contains integral IRs, thus allowing the extraction of various fine-grained cues, the additional object detector inevitably compromises the efficiency of these methods. Furthermore, the cropped proposals lack global contextual information, leading to lower effectiveness. In contrast, the generation of the instance tokens in AGER does not involve an object detector but is optimized as a dynamically clustering process in an end-to-end manner along with the Transformer encoder. Moreover, the clustering mechanism enables instance tokens to be aggregated from a global perception field and potentially eliminates visual redundancy among similar patch tokens, leading to stronger expressiveness of instance tokens.

\vspace{1mm}
\noindent
\textbf{Points} were proposed to represent instances to achieve a one-stage framework for HOI detection. Specifically, \cite{ppdm, ipnet, ggnet} represented the interactions as the midpoints of human-object pairs and detected them based on keypoint detection networks~\cite{newell2016stacked, yu2018deep}, dispensing with additional detectors. Thus, they enjoyed a simpler and faster pipeline, but at the expense of the capability to freely extract extra cues due to the lack of integral IRs.

\vspace{1mm}
\noindent
\textbf{instance queries} were first introduced in the Transformer-based detector~\cite{ detr}, which interact with patch tokens and aggregate information through several interleaved cross-attention and self-attention modules. Thanks to the impressive global context modeling capability, Transformer rapidly revolutionizes HOI detection methods~\cite{endd,dong2021visual,hotr,chen2021reformulating,qpic,zhang2021mining,kim2022mstr,iftekhar2022look,zhou2022human,zhang2022exploring,Iwin,liu2022interactiveness,liao2022gen, tu2022video}. Most works~\cite{endd,qpic,chen2021reformulating} focused on designing an end-to-end pipeline and continuing the success of the attention mechanism for HOI detection, dealing with visual appearance features solely and neglecting the potential of extra instance-level cues. Additionally, some methods~\cite{hotr, liao2022gen} propose to use additional queries to detect instances individually by stacking more decoders. Nevertheless, instance queries are task-driven and fail to extract integral IRs, weakening their ability to extract extra cues due to task bias.  In comparison, our AGER introduces clustering mechanism into Transformer to enable the generation of instance tokens that guarantee the integrality of IRs, which continues the success of global attention and meanwhile enjoys the potential of extra instance-level cues.

\section{Method}
In this section, we aim to explore the solution for a \emph{single-stage} pipeline that allows us to leverage extra instance-level cues for HOI detection. We start with a detailed description of our instance encoder, which incorporates the attention mechanism with dynamical clustering to extract instance tokens, in Sec.~\ref{ins_en}. Then, we take three instance-level cues as guidance to explain the scheme of the extraction and aggregation of extra cues in Sec.~\ref{cue}. Next, in Sec~\ref{int_de}, we propose a new interaction decoder that enumerates all human-object pairs to recognize their interactions in a multi-pattern manner. Finally, we design a special loss function that enables the textual guidance in Sec.~\ref{sec: loss}.

\subsection{Instance Encoder}
\label{ins_en}
As shown in Fig.~\ref{img:whole_model}, the instance encoder is organized as a backbone followed by a hierarchical Transformer encoder, where the latter incorporates self-attention and clustering mechanism to extract instance tokens iteratively.

\vspace{1mm}
\noindent
\textbf{Backbone.} An input image is first downsampled through a plain CNN backbone and then flattened to add a cosine positional embedding to harvest the initialized and sequenced patch tokens $\mathbf{T}_{\text{b}} \in \mathbb{R}^{N_\text{b} \times D_{\text{b}}}$.

\vspace{1mm}
\noindent
\textbf{Overall architecture.} Transformer encoder consists of two stages that share an identical architecture, which comprises of several self-attention layers and a clustering layer. 

Concretely, in the $s$-th stage, we first initialize two sets of learnable clustering centers for humans $\mathbf{C}_{\text{h}}^s \in \mathbb{R}^{N_\text{h}^s \times D_{\text{h}}^s}$  and objects $\mathbf{C}_{\text{o}}^s \in \mathbb{R}^{N_\text{o}^s \times D_{\text{o}}^s}$ separately, which are then 
concatenated with image tokens $\mathbf{T}_{\text{i}}^s$ and learned to update representations through several self-attention (SA) layers. 
Subsequently, at the end of each stage, we assign each image token to different clustering centers based on feature affinities, and the assigned image tokens are then aggregated in the clustering layer. Formally, each stage is computed as
\setlength{\abovedisplayskip}{0.6em}
\setlength{\belowdisplayskip}{0.6em}
\begin{align}
\label{equ:1}
     &[\hat{\mathbf{C}}_{\text{h}}^s ; \hat{\mathbf{C}}_{\text{o}}^s ; \hat{\mathbf{T}}_{\text{i}}^s] = \text{SA-Layer}([\mathbf{C}_{\text{h}}^s ; \mathbf{C}_{\text{o}}^s ; \mathbf{T}_{\text{i}}^s]), \\
     &[\mathbf{G}_{\text{h}}^s ; \mathbf{G}_{\text{o}}^s] = \text{ClusteringLayer}([\hat{\mathbf{C}}_{\text{h}}^s ; \hat{\mathbf{C}}_{\text{o}}^s ; \hat{\mathbf{T}}_{\text{i}}^s]).
\end{align}
Here, $[\: ; \:]$ denotes concatenation operator, $\mathbf{G}_{\text{h}}^s \in \mathbb{R}^{N_\text{h}^s \times D_{\text{h}}^s}$ and $\mathbf{G}_{\text{o}}^s \in \mathbb{R}^{N_\text{o}^s \times D_{\text{o}}^s}$ are the agglomerated image tokens after the $s$-th stage. Note that we omit the modules of token projection, residual connection and normalization here. Specifically, $\mathbf{T}_{\text{i}}^1 = \mathbf{T}_{\text{b}}$ and $\mathbf{T}_{\text{i}}^2 = [\mathbf{G}_{\text{h}}^1 ; \mathbf{G}_{\text{o}}^1]$, \emph{i.e.}, we feed the initialized patch tokens from the backbone to the $1$-th stage, and these small local patches are dynamically agglomerated into relatively larger segments, which are subsequently fed into the $2$-th stage to generate the final instance tokens. Following~\cite{xu2022groupvit}, we propagate the learned clustering centers in the 1st stage to the 2nd stage through a MLP-Mixer layer~\cite{tolstikhin2021mlp}. Meanwhile, to make the human and object clustering centers distinct, we add two sets of position embedding to them. Then, for the human centers, they are obtained via
\begin{align}
\label{euq:2}
    &\mathbf{P}_{\text{h}}^s = \text{Embedding}(N_{\text{h}}^s, D_{\text{h}}^s), \\
    &\tilde{\mathbf{C}}_{\text{h}}^s = \text{Zeros}(N_{\text{h}}^s, D_{\text{h}}^s), \\
    &\mathbf{C}_{\text{h}}^1 = \tilde{\mathbf{C}}_{\text{h}}^1 + \mathbf{P}_{\text{h}}^1, \\
    &\mathbf{C}_{\text{h}}^2 = \tilde{\mathbf{C}}_{\text{h}}^2 + \mathbf{P}_{\text{h}}^2 + \text{MLP-Mixer}(\check{\mathbf{C}}_{\text{h}}^1).
\end{align}
 $\tilde{\mathbf{C}}_{\text{h}}^s$ indicate the centers that are initialized as zeros and $\check{\mathbf{C}}_{\text{h}}^1$ are updated center representations that are calculated by Eq.~\ref{equ:3}. Object centers share the same process.

\vspace{1mm}
\noindent
\textbf{Clustering layer.}
The clustering layer at the end of each stage aims to aggregate local image tokens into a new token based on their feature affinity, thus the small local patch tokens can be gradually merged into a larger segment and finally into an instance token that covers the integral discriminative feature region of an instance.

In particular, we first employ a cross-attention module to update the representation of clustering centers, which enables information propagation between clustering centers and image tokens via
\begin{equation}
\label{equ:3}
\check{\mathbf{C}}_{[\text{h}, \text{o}]}^s = \text{softmax}(\frac{\hat{\mathbf{C}}_{[\text{h}, \text{o}]}^s \cdot (\hat{\mathbf{T}}_{\text{i}}^s)^{\top}}{\sqrt{D_{\text{i}}^s}}) \cdot \hat{\mathbf{T}}_{\text{i}}^s,
\end{equation}
where $\hat{\mathbf{C}}_{[\text{h}, \text{o}]}^s = [\hat{\mathbf{C}}_{\text{h}}^s ; \hat{\mathbf{C}}_{\text{o}}^s]$ is the concatenation of human and object centers from Eq.~\ref{equ:1}. $D_{\text{i}}^s$ is the channel dimension of image tokens. Subsequently, we adopt the scheme in~\cite{xu2022groupvit} to employ a \texttt{Gumbel-softmax}~\cite{jang2017categorical} to compute the similarity matrix $\mathbf{A}^s$ between the clustering centers and the image tokens as
\begin{equation}
    \label{equ:4}
    \setlength{\abovedisplayskip}{1.2em}
\setlength{\belowdisplayskip}{1.2em}
    \mathbf{A}_{(k,j)}^s = \frac{\exp(W_{\text{c}}\check{\mathbf{c}}_k^s \cdot W_{\text{i}}\hat{\mathbf{t}}_j^s + \gamma_i)}{\sum_{n=1}^{N_{\text{c}}^s} \exp(W_{\text{c}}\check{\mathbf{c}}_n^s \cdot W_{\text{i}}\hat{\mathbf{t}}_j^s + \gamma_n)},
\end{equation}
where $\check{\mathbf{c}}_k^s$ stands for the $k$-th clustering center in $\check{\mathbf{C}}_{[\text{h}, \text{o}]}^s$ and $\hat{\mathbf{t}}_j^s$ denotes the $j$-th updated image token in $\hat{\mathbf{T}}_{\text{i}}^s$. $N_{\text{c}}^s = N_{\text{h}}^s + N_{\text{o}}^s$ counts the total number of clustering centers in the $s$-th stage. $W_{\text{c}}$ and $W_{\text{i}}$ are the weights of the learned linear projections for the clustering centers and the image tokens, respectively. $\{\gamma\}_{n=1}^{N_{\text{c}}^s}$ are \emph{i.i.d} random samples drawn from the \texttt{Gumbel(0,1)} distribution that enables the \texttt{Gumbel-softmax} distribution to be close to the real categorical distribution. Finally, we merge $N_{\text{i}}^s$ image tokens with corresponding clustering centers to calculate grouped tokens $\mathbf{G}_{[\text{h},\text{o}]}^s = [\mathbf{G}_{\text{h}}^s ; \mathbf{G}_{\text{o}}^s]$ via
\begin{equation}
    \label{equ:5}
    \mathbf{g}_k^s = \check{\mathbf{c}}_k^s + W_u \frac{\sum_{j=1}^{N_{\text{i}}^s} \mathbf{A}_{(k,j)}^s W_v\hat{\mathbf{t}}_j^s}{\sum_{j=1}^{N_{\text{i}}^s} \mathbf{A}_{(k,j)}^s},
\end{equation}
where $\mathbf{g}_k^s$ is the $k$-th grouped token in $\mathbf{G}_{[\text{h},\text{o}]}^s$, $W_u$ and $W_v$ are learned weights to project the merged features. 

\subsection{Cues Extraction \& Aggregation}

\label{cue}
This work realizes three instance-level cues, \emph{i.e.}, human poses (P), spatial locations (S) and object categories (T), as guidance, other valuable cues can be extracted similarly. 

\vspace{0.8mm}
\noindent
\textbf{Extraction.} Unlike prior methods that use different specially customized models to extract different cues, we extract those cues through several lightweight MLPs in parallel, thanks to the excellent expressiveness of the instance tokens. Concretely, we perform a 5-layer MLP to estimate the normalized locations of 17 keypoints for human pose estimation.  Note that object tokens do not have a pose representation. Meanwhile,  a 3-layer MLP is used to predict the normalized bounding boxes of all humans and objects as their spatial locations. Additionally, we adopt a 1-layer FFN to predict each category of humans and objects $\hat{\mathbf{y}}$. Specifically, for the $i$-th human instance, its prediction $\hat{\mathbf{y}}_{\text{h}}^i \in [0, 1]^2$, where the $2$-th element indicates \emph{no-human}. For object instance, similarly, $ \hat{\mathbf{y}}_{\text{o}}^i \in [0, 1]^{N_{\text{o}}^{\text{c}} + 1}$, where $N_{\text{o}}^{\text{c}}$ is the number of object classes and the $({N_{\text{o}}^{\text{c}} + 1})$-th element denotes \emph{no-object}. 

\vspace{0.8mm}
\noindent
\textbf{Aggregation.} We first adopt two fully connected layers to project all cues into a united and embedded feature space, leading to four new cue representations $\mathbf{E}_{\text{pos}} \in \mathbb{R}^{N_{\text{h}}^{2} \times D_{\text{pos}}}$ (human poses), $\mathbf{E}_{\text{h-spa}} \in \mathbb{R}^{N_{\text{h}}^{2} \times D_{\text{spa}}}$ (human spatial locations), $\mathbf{E}_{\text{o-spa}} \in \mathbb{R}^{N_{\text{o}}^{2} \times D_{\text{spa}}}$ (object spatial locations) and $\mathbf{E}_{\text{cls}} \in \mathbb{R}^{N_{\text{o}}^{2} \times D_{\text{cls}}}$ (object classes).  Particularly, the text of the predicted object name is first transformed into a vector using Word2Vec~\cite{mikolov2013efficient}. Since these cues may introduce noise due to misrecognition, we manually set a threshold $\gamma=0.7$ over the confidence of category prediction to decide their employment. Concretely, if the category (\emph{no-object} and \emph{no-human} are excluded) prediction confidence of an instance is larger than $\gamma$, we keep its corresponding cues otherwise reset them as 0. Finally, these cues are concatenated to corresponding instance tokens to obtain the final representations via:
\begin{align}
    &\hat{\mathbf{T}}_{\text{h}} = W_{\text{h}}[\mathbf{T}_{\text{h}}; \; \mathbf{E}_{\text{pos}}; \; \mathbf{E}_{\text{h-spa}}], \\
    &\hat{\mathbf{T}}_{\text{o}} = W_{\text{o}}[\mathbf{T}_{\text{o}}; \; \mathbf{E}_{\text{cls}}; \; \mathbf{E}_{\text{o-spa}}], 
\end{align}
where $\mathbf{T}_{\text{h}} = \mathbf{G}_{\text{h}}^2$ and $\mathbf{T}_{\text{o}} = \mathbf{G}_{\text{o}}^2$ are human and object tokens generated by the instance encoder. $W_{\text{h}}$ and $W_{\text{o}}$ are learned weights to project the concatenated features.


\subsection{Interaction Decoder}
\label{int_de}
We adopt a 3-layer Transformer decoder to recognize interactions, of which each layer consists of a cross-attention module and a self-attention module. As the clustering mechanism in the instance encoder has located different humans and objects, our decoder aims to associate the interacted human-object pairs and recognize their interactions.

\vspace{1.25mm}
\noindent
\textbf{Association.}
Formally, a given image is invariably transformed into $N_{\text{h}}^{\text{pred}}=N_{\text{h}}^2$ human tokens $\hat{\mathbf{T}}_{\text{h}} \in \mathbb{R}^{N_{\text{h}}^{\text{pred}} \times D_{\text{h}}}$ and $N_{\text{o}}^{\text{pred}}=N_{\text{o}}^2$ object tokens $\hat{\mathbf{T}}_{\text{o}} \in \mathbb{R}^{N_{\text{o}}^{\text{pred}} \times D_{\text{o}}}$ after the instance encoder and the cue utilization module. By design, $D_{\text{h}} = D_{\text{o}}$ and we simplify them as $D$. Then, we add two sets of position embedding to $\hat{\mathbf{T}}_{\text{h}}$ and $\hat{\mathbf{T}}_{\text{o}}$ respectively via:
\begin{align}
    &\mathbf{P}_{\text{h}} = \text{Embedding}(N_{\text{h}}^{\text{pred}}, D), \; \check{\mathbf{T}}_{\text{h}} =\hat{\mathbf{T}}_{\text{h}} + \mathbf{P}_{\text{h}}; \\
    &\mathbf{P}_{\text{o}} = \text{Embedding}(N_{\text{o}}^{\text{pred}}, D), \; \check{\mathbf{T}}_{\text{o}} = \hat{\mathbf{T}}_{\text{o}} + \mathbf{P}_{\text{o}}.
\end{align}
 Next, the position embedding for interaction queries is initialized as the one-to-one sum of the human and object position embedding. Concretely, the position of the $(ij)$-th interaction is the sum of the position of the $i$-th human and the position of the $j$-th object, \emph{ i.e.,} $\mathbf{p}_{\text{a}}^{(ij)} = \mathbf{p}_{\text{h}}^{i} + \mathbf{p}_{\text{o}}^{j}$, leading to an interaction position embedding $\mathbf{P}_{\text{a}} \in \mathbb{R}^{N_{\text{h}}^{\text{pred}}N_{\text{o}}^{\text{pred}} \times D}$, which actually enumerates total $N_{\text{h}}^{\text{pred}}N_{\text{o}}^{\text{pred}}$ possible human-object pairs.
 
 Moreover, in practical scenarios, one human-object pair may have multiple interaction labels. Thus, we follow \cite{wang2022anchor} to incorporate multiple patterns into each interaction position. Concretely, we use a small set pattern embedding $\mathbf{P}_{\text{pattern}} = \texttt{\text{Embedding}}(N_{\text{pattern}}, D)$ to detect different interactions from each human-object pair. $N_{\text{pattern}}$ is the number of patterns that is very small, here $N_{\text{pattern}}=3$. Next, we share the $\mathbf{P}_{\text{pattern}}$ to each interaction position $\mathbf{p}_{\text{a}}$ to get the multi-pattern interaction position embedding $\hat{\mathbf{P}}_{\text{a}} \in \mathbb{R}^{N_{\text{a}} \times D}$, where $N_{\text{a}} = N_{\text{pattern}} \times N_{\text{h}}^{\text{pred}} \times N_{\text{o}}^{\text{pred}}$. Finally, our interaction queries are initialized as:
\begin{equation}
    \mathbf{Q}_{\text{a}} = \text{Zeros}(N_{\text{a}}, D) + \hat{\mathbf{P}}_{\text{a}}.
\end{equation}

\vspace{1.25mm}
\noindent
\textbf{Recognition.} Along with the human and object instance tokens from the instance encoder, we feed the interaction queries $\mathbf{Q}_{\text{a}}$ into the interaction decoder. After that, the interactions are recognized through a 1-layer FFN, following QPIC~\cite{qpic}.

\subsection{Loss Function}
\label{sec: loss}
The loss function consists of three parts: 1) loss of interaction recognition $\mathcal{L}_{\text{a}}$, 2) loss of cues extraction $\mathcal{L}_{\text{e}}$, and 3) loss of instance token generation $\mathcal{L}_{\text{t}}$. Specifically, $\mathcal{L}_{\text{e}}$ consists of pose estimation and location regression. Category recognition is jointly optimized with $\mathcal{L}_{\text{t}}$. We use the focal loss~\cite{lin2017focal} as $\mathcal{L}_{\text{a}}$ and adopt $L_2$ loss as $\mathcal{L}_{\text{e}}$. The total loss is the weighted sum of them, \emph{i.e.}, $\mathcal{L} = \alpha_1\mathcal{L}_{\text{a}} + \alpha_2\mathcal{L}_{\text{e}} + \alpha_3\mathcal{L}_{\text{t}}$. More details are described in \emph{supplementary materials}. Here, we mainly introduce the design of $\mathcal{L}_{\text{t}}$, which enables text representations to guide the generation of instance tokens.

\subsubsection{Textual Guidance} 
Actually, some works have tried to incorporate clustering with Transformer for other tasks, such as GroupViT~\cite{xu2022groupvit} and kMaX~\cite{yu2022k}, and we borrow some ideas from them for model design. However, training the model for HOI detection is not easy. GroupViT use contrastive loss, which demands large training batch size (4096) and kMaX uses heavy decoder and dense annotations.  All of these are unaffordable for HOI detection. Thus, we devise a new loss function that uses a textual signal to guide the learning of the instance encoder by enforcing a similarity between the textual representation and the instance token representation. To this end, we first define a similarity metric and then match instance tokens to each ground truth instance with this metric and finally optimize the instance encoder.

\vspace{1.25mm}
\noindent
\textbf{Similarity metric.} Suppose an input image contains $N_{\text{h}}^{\text{gt}}$ humans and $N_{\text{o}}^{\text{gt}}$ objects, in which the $j$-th object is labeled as $\mathbf{y}_{\text{o}}^j$. Then, taking objects as examples, our similarity metric $\mathrm{sim}(\cdot,\cdot)$ between the $j$-th ground truth object and the $k$-th generated object token $\mathbf{t}_{\text{o}}^k$ is defined as
\begin{equation}
\label{eu:sim}
\text{sim}(j,k) = \hat{\mathbf{y}}_{\text{o}}^k(j) \times \text{Cosine}(\mathbf{r}_{\text{vis}}^k, \mathbf{r}_{\text{txt}}^j),
\end{equation}
where $\hat{\mathbf{y}}_{\text{o}}^k(j) \in [0,1]$ is the probability of predicting the $j$-th class and $\texttt{\text{Cosine}}(\cdot, \cdot)$ denotes cosine similarity. $\mathbf{r}_{\text{vis}}^k$ is visual representation vector projected from the $k$-th object token $\mathbf{t}_{\text{o}}^k$ through two FC layers, and $\mathbf{r}_{\text{txt}}^j$ is a text representation vector from CLIP~\cite{clip}. Concretely, we prompt the \emph{noun} word of $j$-th ground truth class with a handcrafted sentence template, \emph{i.e.}, ``\emph{A photo of a \{noun\}}". Then, we feed this sentence into a frozen text encoder of CLIP followed by two FC layers as projector to acquire the text representation $\mathbf{r}_{\text{txt}}^j$. The human tokens share the same progress. Note that for human, the $j$ ranges from 1 to 2, indicating \emph{human} and \emph{no-human}, while for object, $j = [1, 2, ..., N_{\text{o}}^{\text{c}}, N_{\text{o}}^{\text{c}}+1]$, denoting total $ N_{\text{o}}^{\text{c}}$ different object categories and a \emph{no-object}.

\vspace{1.25mm}
\noindent
\textbf{Instance matching.} By design, $N_{\text{h}}^{\text{pred}} > N_{\text{h}}^{\text{gt}}$ and $N_{\text{o}}^{\text{pred}} > N_{\text{o}}^{\text{gt}}$. We first pad $(N_{\text{h}}^{\text{pred}} - N_{\text{h}}^{\text{gt}})$ and $(N_{\text{o}}^{\text{pred}} - N_{\text{o}}^{\text{gt}})$ ``\emph{nothing}"s to human and object ground truths respectively, leading to two new ground truth sets $\{\mathbf{y}_{\text{h}}^i\}_{i=1}^{N_{\text{h}}^{\text{pred}}}$ and $\{\mathbf{y}_{\text{o}}^j\}_{j=1}^{N_{\text{o}}^{\text{pred}}}$. Following, in case of the object tokens (same for the human tokens), we search for a permutation of $N_{\text{o}}^{\text{pred}}$ elements $\hat{\sigma} \in \mathfrak{S}_{N_{\text{o}}^{\text{pred}}}$  to achieve the maximum total similarity:
\begin{equation}
    \hat{\sigma} = \underset{\sigma \in \mathfrak{S}_{N_{\text{o}}^{\text{pred}}}}{\arg \max} \sum_{i=1}^{N_{\text{o}}^{\text{pred}}}\text{sim}(i, \sigma(i)).
\end{equation}
The optimal assignment is calculated with the Hungarian algorithm~\cite{kuhn1955hungarian}, following DETR~\cite{detr}.

\vspace{1.25mm}
\noindent
\textbf{Objective.} After finding the optimal assignment $\hat{\sigma}$, we are inspired by~\cite{wang2021max} to define the objective taking into account both positive predictions (assigned to ground truths that are not \emph{nothing}) and negative (assigned to \emph{nothing}) predictions into account. In case of object instances, the positive loss is calculated as:
\begin{align}
    \mathcal{L}_{\text{o}}^{\text{pos}} & = \sum_{i=1}^{N_{\text{o}}^{\text{gt}}} \texttt{\text{sg}}(\hat{\mathbf{y}}_{\text{o}}^{\hat{\sigma}(i)}(i)) \cdot [- \text{Cosine}(\mathbf{r}_{\text{vis}}^{\hat{\sigma}(i)}, \mathbf{r}_{\text{txt}}^i)]  \notag \\ 
    & + \sum_{i=1}^{N_{\text{o}}^{\text{gt}}} \texttt{\text{sg}}(\text{Cosine}(\mathbf{r}_{\text{vis}}^{\hat{\sigma}(i)}, \mathbf{r}_{\text{txt}}^i)) \cdot [-\log \hat{\mathbf{y}}_{\text{o}}^{\hat{\sigma}(i)}(i)].
\end{align}
Intuitively, $\mathcal{L}_{\text{o}}^{\text{pos}}$ is equivalent to optimizing a cosine loss weighted by the class correctness and optimizing a cross-entropy loss weighted by the cosine similarity. Note that the stop gradient operator \texttt{sg}$(\cdot)$ ensures constant loss weights. If a token is mis-recognized, we disregard its representation since it is a false negative anyway. The wrong representation also downscales the weight of the recognition loss. Thus, we enforce the representation and class to be correct at the same time. Besides, we define the negative loss as:
\begin{equation}
    \mathcal{L}_{\text{o}}^{\text{neg}} = \sum_{j=N_{\text{o}}^{\text{gt}}+1}^{N_{\text{o}}^{\text{pred}}}[-\log\hat{\mathbf{y}}_{\text{o}}^{\hat{\sigma}(j)}(N_{\text{o}}^{\text{c}} + 1)].
\end{equation}
 Finally, the objective for the object instances is designed as $\mathcal{L}_{\text{t}}^{\text{o}} = \lambda \mathcal{L}_{\text{o}}^{\text{pos}} + (1-\lambda)\mathcal{L}_{\text{o}}^{\text{neg}}$. The objective of human instances $\mathcal{L}_{\text{t}}^{\text{h}}$ shares the same process. Finally, $\mathcal{L}_{\text{t}} = \mathcal{L}_{\text{t}}^{\text{o}} + \mathcal{L}_{\text{t}}^{\text{h}}$.

\section{Experiments}

\begin{table*}[t]
\vspace{-1.2em}
    \centering
\subfloat[{\footnotesize \textbf{Extra cues improve HOI detection}}
\label{tab:cues_per}
]{
\centering
\begin{minipage}{0.3\linewidth}{\begin{center}
\tablestyle{4pt}{1.3}
\begin{tabular}{y{20}|x{50}x{53}}
Cues & Full & Rare \\
\shline
\baseline{A} & \baseline{32.15} & \baseline{23.81} \\
A+P & 33.79 {\color{deltacolor}{(+$5.1\%$)}} & 25.63 {\color{deltacolor}{(+$7.6\%$)}}\\
A+S & 32.74 {\color{deltacolor}{(+$2.0\%$)}} & 24.61 {\color{deltacolor}{(+$3.4\%$)}}\\
A+T & 34.08 {\color{deltacolor}{(+$6.0\%$)}} & 26.21 {\color{deltacolor}{(+$10.1\%$)}}\\
\end{tabular}
\end{center}}\end{minipage}
}
\hspace{1em}
\subfloat[
{\footnotesize \textbf{mAP gain varies with different sample volumes}}
\label{tab:num}
]{
\begin{minipage}{0.39\linewidth}{\begin{center}
\tablestyle{4pt}{1.3}
 \begin{tabular}{x{50}|x{40}x{40}x{40}}
    Number & w/o   &  w/ & $\Delta \: \uparrow$ \\
    \shline
    5000/500 & 61.42/15.37 & 69.83/27.52 & 8.41/\textbf{12.15} \\
    5000/1000 & 61.38/20.21 & 69.86/28.64 & \textbf{8.48}/8.43 \\
    5000/3000 & 58.63/39.07 & 65.74/43.86  & 7.11/4.79 \\ 
    \baseline{5000/5000} & \baseline{57.45/52.51} & \baseline{61.14/55.28} & \baseline{3.69/2.77}
    \end{tabular}
    \end{center}
    }\end{minipage}
    }
\hspace{1em}
\subfloat[
{\footnotesize \textbf{Cues differ in recognizability}}
\label{tab:diff}
]{
\begin{minipage}{0.22\linewidth}{\begin{center}
\tablestyle{1pt}{1.3}
\begin{tabular}{x{30}|x{24}}
Feature & Diff. \\
\shline
\baseline{A} & \baseline{0.06}  \\
P & 0.13 \\
S & 0.08 \\
T & - \\
\end{tabular}
\end{center}}\end{minipage}
}
    
\captionsetup{size=small}
\caption{\textbf{Importance of instance-level cues}. (a) The results of incorporating visual appearance features (A) with other cues in Sec.~\ref{cue}. (b) The results of using (w/) and not using (w/o) extra cues with different sample volumes. (c) The mean differences of different cues.}
\end{table*}

\begin{figure*}[t]
    \centering
     \subfloat[
{\footnotesize \textbf{Coverage}}
\label{fig:cov}
]{
    \begin{minipage}[t]{0.32\linewidth}
    \centering
    \includegraphics[width=0.97\linewidth]{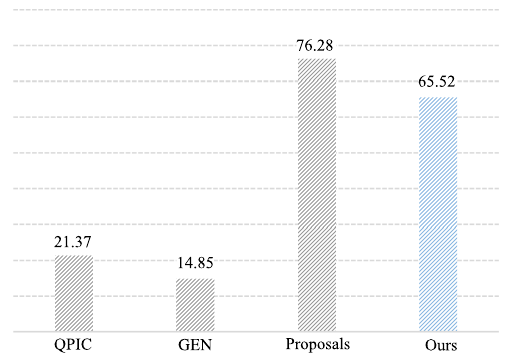}
    \end{minipage}
    }
    \subfloat[
{\footnotesize\textbf{Accuracy}}
\label{fig:acc}
]{
    \begin{minipage}[t]{0.32\linewidth}
    \centering
    \includegraphics[width=0.97\linewidth]{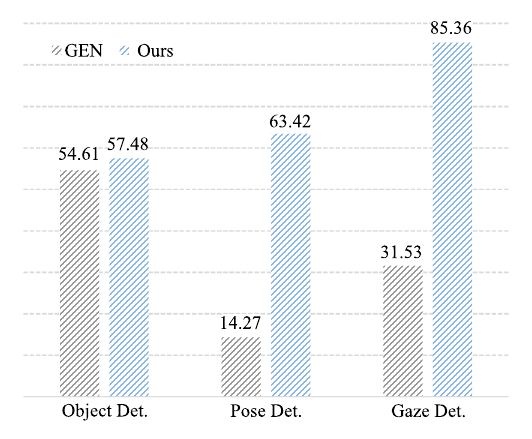}
    \end{minipage}
    }
   \subfloat[
{\footnotesize\textbf{Model size}} 
\label{fig:m_size}
]{
    \begin{minipage}[t]{0.32\linewidth}
    \centering
    \includegraphics[width=0.97\linewidth]{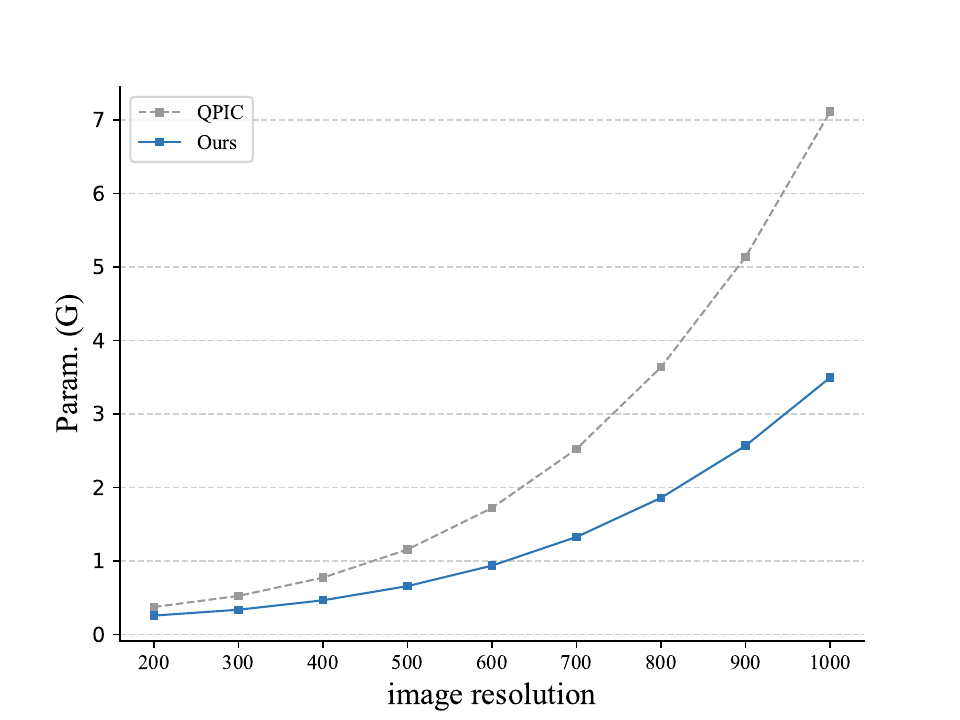}
    \end{minipage}
    }
    \captionsetup{size=small}
    \caption{\textbf{Importance of clustering.} (a) We report the coverage as the proportion of the area of the feature region highlighted by an information carriers to the area of an instance. (b) Performance of different information carriers for different tasks (both fine-tuned with the same supervisory signals as ours). (c) The model parameters over image resolution.}
    \label{fig:cues}
\end{figure*}

\noindent
\textbf{Technical details.} Most of our default settings follow QPIC~\cite{qpic}, \emph{e.g.}, data augmentation, backbone, etc.
Specifically, the channel dimension of all tokens, clustering centers and position embedding are set to 256. $D_{\text{pos}} = 64$, $D_{\text{spa}} = 16$, $D_{\text{cls}} = 64$. We design $N_{\text{h}}^1 = 16; N_{\text{h}}^2=4$ and $N_{\text{o}}^1 = 64; N_{\text{o}}^2=8$. There are 4 and 2 self-attention layers in the first and second stage. For loss calculation, $\alpha_{1,2,3}$ are set to 2.5, 1, 1.5, and $\lambda=0.75$. For human pose, we use the annotations provided by~\cite{fang2017rmpe,li2020pastanet} for HICO-Det~\cite{relate1} and the annotations from~\cite{jin2020whole} for V-COCO~\cite{gupta2015visual}.

\vspace{1mm}
\noindent
\textbf{Training.} Our batch size is 32, with an initialized learning rate of the backbone $10^{-5}$, that of the others $2.5e^{-4}$, and the weight decaye $10^{-4}$. We adopt the AdamW~\cite{adamw} optimizer for a total of 150 epochs where learning rates are decayed after 80 and 120 epochs.

\begin{table*}[t]
    \centering
    \tablestyle{4pt}{0.5}
    \begin{tabular}{y{45}x{40}x{40}|x{20}x{30}|x{30}x{30}x{35}|x{30}x{30}x{35}}
    \shline
    & & & & & \multicolumn{3}{c|}{\small Default} & \multicolumn{3}{c}{\small Know Object} \\
    Method & Detector & Backbone & Cues & Single & Full & Rare & Non-Rare & Full & Rare & Non-Rare \\
    \shline
    \multicolumn{2}{l}{\scriptsize CNN-based Methods:}  & & & & & & & & \\
    InteractNet~\cite{interactions} & COCO & R50-FPN &  \XSolidBrush &  \XSolidBrush & 9.94 & 7.16 & 10.77 &- &-&- \\
    iCAN~\cite{ican} & COCO & R50 & \XSolidBrush & \XSolidBrush & 14.84 & 10.45 & 16.15 & 16.26 & 11.33 & 17.73 \\
    PMFNet~\cite{wan2019pose} & COCO & R50-FPN & \Checkmark & \XSolidBrush & 17.46 & 15.65 & 18.00 & 20.34 &17.47 &21.20\\
    DRG~\cite{drg} & COCO & R50-FPN & \Checkmark & \XSolidBrush &19.26 &17.74 &19.71 &23.40 &21.75 &23.89\\
    FCMNet~\cite{liu2020amplifying} & COCO & R50 & \Checkmark & \XSolidBrush &20.41 &17.34 &21.56 &22.04 &18.97 &23.12\\
    DJ-RN~\cite{li2020detailed} & COCO & R50 & \Checkmark & \XSolidBrush &21.34 &18.53 &22.18 &23.69 &20.64 &24.60\\
    SCG~\cite{zhang2021spatially} & COCO & R50-FPN & \Checkmark & \XSolidBrush &21.85 &18.11 &22.97 &- &-&- \\
    UnionDet~\cite{uniondet} & COCO & R50 & \XSolidBrush & \Checkmark &17.58 &11.72 &19.33 &19.76 &14.68 &21.27\\
    IP-Net~\cite{ipnet} & COCO &Hg-104 & \XSolidBrush & \Checkmark &19.56 &12.79 &21.58 &22.05 &15.77 &23.92\\
    PPDM~\cite{ppdm} & HICO-Det &Hg-104 & \XSolidBrush & \Checkmark &21.94 &13.97 &24.32 &24.81 &17.09 &27.12\\
    GG-Net~\cite{ggnet} & HICO-Det &Hg-104 & \XSolidBrush & \Checkmark &23.47 &16.48 &25.60 &27.36 &20.23 &29.48\\
    \shline
    \multicolumn{2}{l}{\scriptsize Transformer-based Methods:}  & & & & & & & & \\
    HOI-T~\cite{endd}  & HICO-Det & R50 & \XSolidBrush & \Checkmark &23.46 &16.91 &25.41 &26.15 &19.24 &28.22\\
    PST~\cite{dong2021visual} & - & R50 & \XSolidBrush & \Checkmark &23.93 &14.98 &26.60 &26.42 &17.61 &29.05\\
    HOTR~\cite{hotr} & HICO-Det & R50 & \XSolidBrush & \Checkmark &25.10 &17.34 &27.42 &- &- &-\\
    AS-Net~\cite{chen2021reformulating} & HICO-Det & R50 & \XSolidBrush & \Checkmark &28.87 &24.25 &30.25 &31.74 &27.07 &33.14\\
    QPIC~\cite{qpic} & HICO-Det & R101 & \XSolidBrush & \Checkmark &29.90 &23.92 &31.69 &32.38 &26.06 &34.27\\
    CDN-L~\cite{zhang2021mining} & HICO-Det & R101 & \XSolidBrush & \Checkmark &32.07 &27.19 &33.53 &34.79 &29.48 &36.38\\
    MSTR~\cite{kim2022mstr} & HICO-Det & R50 & \XSolidBrush & \Checkmark &31.17 &25.31 &32.92 &34.02 &28.83 &35.57\\
    SSRT~\cite{iftekhar2022look} & HICO-Det & R50 & \XSolidBrush & \Checkmark &31.34 &24.31 &33.32 &- &-&- \\
    DT~\cite{zhou2022human} & HICO-Det & R50 & \XSolidBrush & \Checkmark &31.75 &27.45 &33.03 &34.50 &30.13 &35.81\\
    STIP~\cite{zhang2022exploring} & HICO-Det & R50 & \Checkmark & \XSolidBrush &32.22 &28.15 &33.43 &35.29 &31.43 &36.45\\
    Iwin~\cite{Iwin} & HICO-Det & R101 & \XSolidBrush & \Checkmark &32.79 &27.84 &35.40 &35.84 &28.74 &36.09\\
    IF~\cite{liu2022interactiveness} & HICO-Det & R50 & \XSolidBrush & \Checkmark &33.51 &30.30 &34.46 &36.28 &33.16 &37.21\\
    GEN~\cite{liao2022gen}& HICO-Det & R101 & \XSolidBrush & \Checkmark &34.95 &31.18 &36.08 &38.22 &34.36 &39.37\\
    \baseline{Our w/o cues } & \baseline{HICO-Det} & \baseline{R50} & \baseline{\Checkmark} & \baseline{\Checkmark}  & \baseline{33.07} &\baseline{29.87} &\baseline{34.05} &\baseline{35.21} &\baseline{32.04} &\baseline{37.09}\\
    \baseline{Our w/ cues } & \baseline{HICO-Det} & \baseline{R50} & \baseline{\Checkmark} & \baseline{\Checkmark}  & \baseline{\textbf{36.75}} &\baseline{\textbf{33.53}} &\baseline{\textbf{37.71}} &\baseline{\textbf{39.84}} &\baseline{\textbf{35.58}} &\baseline{\textbf{40.23}}\\
    \shline
    \end{tabular}
    \captionsetup{size=small}
    \caption{\textbf{Performance comparison on the HICO-Det test set}. We present an additional tag ``Cues" to indicate the ability to flexibly use a variety of instance-level cues, as well as ``Single" to denote a single-stage pipeline.}
    \label{tab:sota}
\end{table*}
\begin{table}[t]
    \centering
    \tablestyle{4pt}{1}
    \begin{tabular}{y{50}x{20}|x{40}x{40}}
    Method & Cues & $\text{AP}_{\text{role}}^{\text{S1}}$  & $\text{AP}_{\text{role}}^{\text{S2}}$\\
    \shline
    iCAN~\cite{ican} & \XSolidBrush & 45.03 &52.40 \\
    FCMNet~\cite{liu2020amplifying} & \Checkmark &53.10 &-\\
    AS-Net~\cite{chen2021reformulating} & \XSolidBrush & 53.90 & - \\
    QPIC~\cite{qpic} & \XSolidBrush & 58.80 & 60.00 \\
    Iwin~\cite{Iwin} & \XSolidBrush & 60.85 &- \\
    STIP~\cite{zhang2022exploring} & \Checkmark & \textbf{66.00} & \textbf{70.70} \\
    GEN~\cite{liao2022gen} &\XSolidBrush & 63.58 &65.93 \\
    
    \baseline{Ours} & \baseline{\Checkmark} & \baseline{65.68} & \baseline{69.72}
    \end{tabular}
    \vspace{-0.5em}
    \captionsetup{size=small}
    \caption{\textbf{Performance  on the V-COCO}. Limited by space, the detailed comparison is listed in \emph{supplementary materials}.}
    \label{tab:vcoco}
\end{table}
\subsection{Importance of Instance-level Cues}

This subsection aims to verify the importance of different instance-level cues and explore why they facilitate interaction recognition. As Tab.~\ref{tab:cues_per} verified, all cues contribute a performance gain for HOI detection, especially for the ``rare" case (with fewer than 10 training instances), ranging from $3.4\%$ to $10.1\%$. The transformer shows excellent performance when dealing with a large number of training samples yet an inferior performance with inadequate sample volume due to the lack of inductive bias~\cite{vit}. However, HOI detection has always been plagued by the long-tail distribution problem, interactions (\emph{e.g.}, \emph{stand on chair}) with a minority of samples are thereby more likely to be misrecognized as an interaction (\emph{e.g.}, \emph{sit on chair}) with similar visual pattern but massive samples. In this case, instance-level cues serve as some explicit priori knowledge that may be prioritised by the Transformer to recognize interactions. We further verify this solution in Tab.~\ref{tab:num}. Concretely, we choose 5,000 images of \emph{wheel} bicycle and 5,000 images of \emph{ride} bicycle to retrain the interaction decoder with the instance encoder being frozen. As the table shows, when the sample volumes of two interaction instances differ substantially (\emph{e.g.}, 500 vs. 5,000), additional cues can significantly improve performance, especially for small samples ($79.1\%$ vs. $13.7\%$). However, the gain diminishes as the sample size tends to equalize ($5.3\%$ vs. $6.4\%$ with 5,000/5,000 samples). Additionally, Tab.~\ref{tab:diff} reports the mean difference between the cues extracted from these two interaction examples. Empirically, a relatively larger mean difference indicates a better recognizability and thus facilitates the process of classification. From this point, the various instance-level cues are more desirable features for interaction recognition.

\subsection{Importance of Clustering}
\label{sec:clustering}
As mentioned previously, the integrality of IRs is the cornerstone of extracting different cues in a single-stage framework. Fig.~\ref{fig:cov} first shows the coverage rate of different instance information carriers over the instance bounding box. Concretely, the proposals extracted by an extra object detector (DETR in here) show best performance, but enforces a two-stage pipeline that compromises the efficiency. Meanwhile, object detection-driven instance queries in GEN~\cite{liao2022gen} attend to instance parts (14.85\% coverage rate), which leads to an inferior performance in extracting other cues, as shown in Fig.~\ref{fig:acc}. In comparison, the instance tokens generated by clustering enable a sufficient coverage over instances, allowing one to flexibly extract different extra cues ($3\times$ precision improvement). Interestingly, the clustering mechanism natively eliminates the visual redundancy in similar tokens, promising the instance tokens a capability for increased expressiveness. Therefore, even without using any additional decoder, AGER already shows a competitive result of object detection (57.48@AP50) compared to other more complex methods.

\subsection{Analysis of Effectiveness \& Efficiency}
\label{sec:eff}
\noindent
\textbf{Effectiveness}. Tab.~\ref{tab:sota} and Tab.~\ref{tab:vcoco} verify the effectiveness of AGER on HICO-Det~\cite{relate1} and V-COCO~\cite{gupta2015visual}, respectively. First, AGER even without involving any additional cues already achieves a competitive result, with a relative $10.6\%$ mAP gain compared to QPIC~\cite{qpic} on HICO-Det. It is ascribable to the CLIP-guided dynamic clustering process, which reduces the visual redundancy in patch tokens and leads to more expressive instance tokens.  Secondly, AGER achieves a new state-of-the-art performance (36.75 mAP) based on human poses, spatial distributions and object categories.  Note that this result can be further improved by using more valuable cues (37.10/37.77 mAP with gaze/interactiveness) at a negligible cost of additional parameters (+2.36M). However, we are not striving for that, but aim to provide the first paradigm that enables us to use extra cues in a single-stage manner, giving some valuable points to the HOI detection community. Although AGER does not achieve the optimal results on V-COCO, its performance is still very competitive. 

\begin{table}[t]
    \centering
    \tablestyle{4pt}{1}
    \begin{tabular}{y{45}|x{40}x{40}x{20}}
    
    Method & Param. & GFlOPs & FPS\\
    \shline
    QPIC~\cite{qpic} & \textbf{42.35M} & 36.95 & 20.0 \\
    AS-Net~\cite{chen2021reformulating} & 59.14M & 52.94 & 1.6 \\
    STIP~\cite{zhang2022exploring} & 54.71M & 48.27 & 1.6 \\
    \baseline{Ours} & \baseline{44.47M} & \baseline{\textbf{33.81}} & \baseline{\textbf{27.2}}
    \end{tabular}
    \captionsetup{size=small}
    \caption{\textbf{Analysis of efficiency}. All models are tested using a sigle GTX 1080Ti taking as input an image with a size of $640 \times 640$. Here, we adopt ResNet50-FPN as the backbone. }
    \label{tab:eff}
\end{table}
\begin{table*}[t]
    \centering
\subfloat[
\textbf{Centers number}. 
\label{tab:cen_num}
]{
\centering
\begin{minipage}{0.24\linewidth}{\begin{center}
\tablestyle{2pt}{1.05}
\begin{tabular}{x{28}x{28}|x{24}x{24}}
Stage 1 & Stage 2 & Full &Rare \\
\shline
(32, 32) & (4, 4)& 30.19 & 26.24\\
\baseline{(16, 64)} & \baseline{(4, 8)} & \baseline{\textbf{36.75}} & \baseline{\textbf{33.53}} \\
(32, 64) & (8, 8) & 33.31 & 31.07 \\
(32, 64) & (8, 16) & 34.42 & 31.93\\
\end{tabular}
\end{center}}\end{minipage}
}
\hspace{0.2em}
\subfloat[
\textbf{Patterns number}. 
\label{tab:pat-num}
]{
\begin{minipage}{0.23\linewidth}{\begin{center}
\tablestyle{2pt}{1.05}
\begin{tabular}{x{26}|x{24}x{24}}
Pattern & Full & Rare\\
\shline
1 & 34.81 & 29.40 \\
\baseline{3} & \baseline{\textbf{36.75}} & \baseline{\textbf{33.53}} \\
5 & 35.54 &  32.89\\
7 & 35.10 & 32.81\\
\end{tabular}
\end{center}}\end{minipage}
}
\hspace{0.2em}
\subfloat[
\textbf{Training strategies}.  
\label{tab:str}
]{
\begin{minipage}{0.22\linewidth}{\begin{center}
\tablestyle{2pt}{1.05}
 \begin{tabular}{x{44}|x{24}x{24}}
    Strategy & Full & Rare\\
    \shline
    \baseline{Base} & \baseline{\textbf{36.75}} & \baseline{\textbf{33.53}}\\
   - Center pos. & 33.71 & 30.26\\
   - Cue-Switch & 34.26 & 29.82\\
    - CLIP & 19.82 & 12.53
    \end{tabular}
    \end{center}
    }\end{minipage}
    } 
\hspace{0.2em}
\subfloat[
\textbf{Similarity metric}. 
\label{tab:s-met}
]{
\begin{minipage}{0.21\linewidth}{\begin{center}
\tablestyle{2pt}{1.05}
\begin{tabular}{x{40}|x{24}x{24}}
metric & Full & Rare\\
\shline
CE & 19.82   &  14.53\\
Cos & 29.80  & 25.32\\
CE+Cos & 33.21 & 29.40\\
\baseline{weighted} & \baseline{\textbf{36.75}} & \baseline{\textbf{33.53}}
\end{tabular}
\end{center}}\end{minipage}
}
\vspace{-0.6em}
\captionsetup{size=small}
\caption{\textbf{Ablations}. In (a), $(\cdot,\cdot)$ denotes (human,object); In (c) ``-" means ``w/o". All experiments are conducted on HICO-Det test set.}
\end{table*}

\vspace{0.8mm}
\noindent
\textbf{Efficiency.} In Tab.~\ref{tab:eff}, we compare four different yet typical Transformer-based methods, including: (\textbf{i}) QPIC~\cite{qpic} that adopt a vanilla DETR-like Transformer (6-layer encoder and 6-layer decoder); (\textbf{ii}) AS-Net~\cite{chen2021reformulating} that performs two decoders to detect instances and interactions respectively (6-layer encoder and $2 \times 6$-layer decoder); (\textbf{iii}) STIP~\cite{zhang2022exploring} that built on a two-stage pipeline where instances are first detected through DETR~\cite{detr} and (\textbf{iv}) our AGER. As shown in the table, AGER is even more efficient than QPIC that has the most simple architecture in prior Transformer-based HOI detection methods, with a relative $36.0\%$ gain of FPS and a $8.5\%$ reduction of FLOPs. Formally, additional computational costs of ATER are mainly introduced by calculating instance-level cues and clustering centers. However, for the former, thanks to the expressiveness of instance tokens, several lightweight MLPs are adequate to extract different cues, which bring a negligible additional complexity compared to the method using different customized tools. For the latter, although the first stage of the instance encoder takes more computation to update the clustering centers, the second stage starts to process much less tokens after clustering, and the number of tokens is further reduced after the second stage. Thus, the decoder demands a minority of computational complexity. Meanwhile, thanks to the great representation ability of the instance tokens, the decoder of AGER is much shallower than that of QPIC (3 vs. 6). Also, unlike QPIC has a quadratic computational cost \emph{w.r.t} the number of pixels, the size of input image does not introduce additional computations to AGER but the first stage of encoder. This is because except the first stage, AGER deals with a fixed number of tokens regardless of the input size. We visualize the relations between the complexity of different methods and the image resolution in Fig.~\ref{fig:m_size}, and present a detailed validation in \emph{supplementary materials}.

\subsection{Ablation Study}
\noindent
\textbf{Clustering center numbers.} In Tab.~\ref{tab:cen_num}, we compare different numbers of clustering centers. Overall, increasing centers consistently improves performance, and we find (16,64) for the first stage and (4,8) for the second stage to be optimal. Empirically, an inadequate amount of centers may fail to characterize an image sufficiently, while an excessive amount of centers are likely to introduce unexpected noises.

\vspace{0.8mm}
\noindent
\textbf{Pattern numbers.} Tab.~\ref{tab:pat-num} shows the effect of multi-pattern mechanism in the interaction decoder. Specially, when the number of patterns is one, we adopt the strategy of QPIC~\cite{qpic} to predict a \emph{not-one-hot-like} label, ~\emph{i.e.}, a label with multiple true values. However, such an intuitive solution brings more ambiguity. In contrast, our multi-pattern strategy explicitly encourages each position embedding to attend to one specific interaction, leading to a relative $5.6\%$ mAP gain.

\vspace{.8mm}
\noindent
\textbf{Strategies.} We verify the effectiveness of the proposed strategies in Tab.~\ref{tab:str}. Concretely, without explicitly adding position embedding to human and object centers respectively, the increased ambiguity leads to a $8.3\%$ performance degradation. Besides, we observe a relative $6.8\%$ degradation when invalidating the ``\emph{cue-switch}" strategy in cue aggregation module (Sec.~\ref{cue}), \emph{i.e.}, treating all generated instance tokens as valid without using the threshold $\gamma$ to invalidate mis-recognized instances. Note that our utilization of CLIP is quite different from other methods. Concretely, other methods (\emph{e.g.}, GEN~\cite{liao2022gen}) perform CLIP to transfer interaction-specific linguistic knowledge to a visual model by using interaction (HOI-specific) labels to customize an interaction classifier, while we use just instance labels to generate general IRs. Actually, the majority of HOI detection methods use such general IRs since they are initialized using a pre-trained object detection or classification network.

\vspace{0.8mm}
\noindent
\textbf{Similarity metric.} Tab.~\ref{tab:s-met} compares different similarity metrics for our new objective function.   When using cross-entropy (CE) solely, \emph{i.e.}, involving no textural guidance, we observe severe performance degradation ($\approx 50\%$), indicating that simple CE loss cannot enable dynamical clustering. We conjecture that using CE loss is more like a recognition task that may introduce unexpected task bias, \emph{i.e.}, highlighting partial features. In comparison, text representation is decoupled from downstream tasks and thus involves no task-bias. However, when adopting cosine similarity individually, we also observe a $18.9\%$ performance degradation. It is because that the frozen text encoder of CLIP cannot differentiate two instances in the same category but with different attributes (\emph{e.g.}, a standing \underline{human} and a sitting \underline{human}) as they are both labeled as ``\emph{a photo of a human}". If we jointly train the text encoder and provide more fine-grained labels (\emph{e.g.}, \emph{a photo of a standing human}), the results should be improved, yet introduce much more training complexity and annotation workload. In comparison, our proposed loss is a dynamical fusion of features' generality (both human) and variability (with different attributes), which eliminates task-bias and also facilitates model training.

\section{Discussion \& Conclusion}
\noindent
\textbf{Limitation.} We find that clustering demands a relative higher resolution, so AGER struggles to handle small and occluded instances. Besides, our instance decoder enumerates all human-object pairs without considering interactiveness. All of these await further exploration.

\vspace{.8mm}
\noindent
\textbf{Conclusion.} In this paper, we present AGER, a novel vision Transformer for HOI detection, which provides the first paradigm that enables Transformer-based HOI detector to leverage extra cues in an efficient (single-stage) manner. AGER performs tokenization as a text-guided dynamic clustering process, improving prior methods with instance tokens, which ensures the integrality of IRs. We validate AGER on two challenging HOI benchmarks and achieve a considerable performance boost over SOTA results.

{\small
\bibliographystyle{ieee_fullname}
\bibliography{egbib}
}

\empty

\section{Appendix}

\textbf{Datasets.}
 We conducted experiments on HICO-Det~\cite{relate1} and V-COCO~\cite{gupta2015visual} benchmarks to evaluate the proposed method by following the standard scheme. Specifically, HICO-Det contains 38,118 and 9,658 images for training and testing, and includes 600 HOI categories(\emph{full}) over 117 interactions and 80 objects. It was further split into 138 Rare (with less than 10 training instances) and the other 462 None-Rare categories. V-COCO is a relatively smaller dataset that originates from the MS COCO~\cite{coco}. It consists of 2,533 and 2,867 images for training, validation, as well as 4,946 ones for testing. The images are annotated with 80 object and 29 action classes.

\vspace{1.25mm}
\noindent
\textbf{Evaluation metrics.} We adopt the commonly utilized mean average precision (mAP) to evaluate model performance on both datasets. A predicted HOI instance is considered as true positive if and only if the predicted human and object bounding boxes both have IoUs larger than 0.5 with the corresponding ground truth bounding boxes, and the predicted action label is correct.

Moreover, for HICO-Det, we evaluate model performance in two different settings following~\cite{relate1}: (1) \textbf{Known-object}. For each HOI category, we evaluate the detection only on the images containing the target object category. (2) \textbf{Default}. For each HOI category, we evaluate the detection on the full testset, including images that may not contain the target object. For V-COCO, we report the role mAPs for two scenarios: S1 for the 29 action categories including 4 body motions and S2 for the 25 action categories without the no-object HOI categories.

\section{Loss Function}
\label{sec: 2}

As mentioned in the main text, our loss function consists of three parts and is defined as $\mathcal{L} = \alpha_1\mathcal{L}_{\text{a}} + \alpha_2\mathcal{L}_{\text{e}} + \alpha_3\mathcal{L}_{\text{t}}$. We report different performances of different loss-weighted combinations in Table~\ref{tab:loss}.

\begin{table}[t]
    \centering
    \tablestyle{4pt}{1}
    \begin{tabular}{x{15}x{15}x{15}|x{40}x{40}}
    $\alpha_1$ & $\alpha_2$ & $\alpha_3$  & Full & Rare\\
    \shline
    1 & 1 & 1 & 34.13 &31.20 \\
    2 & 1 & 1 & 35.68 & 34.71\\
    1 & 2 & 1 & 33.97 & 30.84\\
    1 & 1 & 2 & 35.19 & 34.26\\
    \baseline{2.5} & \baseline{1} & \baseline{1.5} & \baseline{\textbf{36.75}} & \baseline{\textbf{33.53}}\\
    \end{tabular}
    \captionsetup{size=small}
    \caption{\textbf{Loss weights}. Performances of different loss-weighted combinations.}
    \label{tab:loss}
\end{table}
\begin{table}[t]
    \centering
    \tablestyle{4pt}{1}
    \begin{tabular}{y{50}x{20}|x{40}x{40}}
    Method & Cues & $\text{AP}_{\text{role}}^{\text{S1}}$  & $\text{AP}_{\text{role}}^{\text{S2}}$\\
    \shline
    InteractNet~\cite{interactions} & \XSolidBrush &40.0 &- \\
    iCAN~\cite{ican} & \XSolidBrush & 45.03 &52.40 \\
    DRG~\cite{drg} & \Checkmark &51.0 &- \\
    IP-Net~\cite{ipnet} & \XSolidBrush & 51.0 &- \\
     PMFNet~\cite{wan2019pose} & \Checkmark &52.0 &- \\
    FCMNet~\cite{liu2020amplifying} & \Checkmark &53.10 &-\\
    GG-Net~\cite{ggnet} & \XSolidBrush &54.7 &-\\
    AS-Net~\cite{chen2021reformulating} & \XSolidBrush & 53.90 & - \\
    HOTR~\cite{hotr} & \XSolidBrush & 55.2 &64.4\\
    QPIC~\cite{qpic} & \XSolidBrush & 58.80 & 60.00 \\
    Iwin~\cite{Iwin} & \XSolidBrush & 60.85 &- \\
    STIP~\cite{zhang2022exploring} & \Checkmark & \textbf{66.00} & \textbf{70.70} \\
    GEN~\cite{liao2022gen} &\XSolidBrush & 63.58 &65.93 \\
    
    \baseline{Ours} & \baseline{\Checkmark} & \baseline{65.68} & \baseline{69.72}
    \end{tabular}
    \captionsetup{size=small}
    \caption{\textbf{Performance  on the V-COCO}. }
    \label{tab:coco}
\end{table}
\section{Result on V-COCO}
\label{coco}
We report a more comprehensive results on the V-COCO in the Table~\ref{tab:coco}.

\section{Ablation Studies.}
\label{sec:as}

\noindent
\textbf{Efficiency.} The computational complexities of Transformer are most introduced by the calculation of attention weights, including self-attention (SA) and cross-attention (CA). Given an image, suppose there are $N$ tokens after backbone, and each token's dimension is $C$. For QPIC~\cite{qpic} and our AGER, the computational complexities are:
\begin{align}
    \Omega(\text{QPIC}) = &\underbrace{6(4NC^2 + 2N^2C)}_{\text{6-layer encoder (SA)}} + \underbrace{6(4N_{\text{q}}C^2 + 2N_{\text{q}}^2C)}_{\text{6-layer decoder (SA)}} \notag \\
    & +\underbrace{6(2N_{\text{q}}C^2 + 2NC^2 + NN_{\text{q}}C + N^2C)}_{\text{6-layer decoder (CA)}},
\end{align}

\begin{align}
    \Omega(\text{AGER}) = &\underbrace{4[4(N+64+16)C^2 + 2(N+80)^2C]}_{\text{first stage (4-layer SA)}}  \notag \\ 
    &+\underbrace{2[4(80+8+4)C^2 + 2(92)^2C]}_{\text{second stage (2-layerSA)}} \notag \\
    &+\underbrace{3(4N_{\text{q}}C^2 + 2N_{\text{q}}^2C)}_{\text{3-layer decoder (SA)}} \notag \\
    & +\underbrace{3(2N_{\text{q}}C^2 + 2\cdot12C^2 + 12N_{\text{q}}C + 12^2C)}_{\text{3-layer decoder (CA)}},
\end{align}
where $N_{\text{q}}$ is the number of additional query embedding inputted into decoder. Concretely, in QPIC, $N_{\text{q}} = 100$ while $N_{\text{q}} = 3 \times (4+8) = 36$ in AGER, where $3$ is the number of patterns and $4+8=12$ is the number of total instance tokens. $C=256$. By calculating $\Omega(\text{QPIC}) - \Omega(\text{AGER}) > 0$, we have $n=-71$. Namely, For arbitrary image, AGER has a lower complexity than QPIC.

\end{document}